\documentclass[conference]{IEEEtran}
\usepackage[linesnumbered,ruled,vlined]{algorithm2e} 
\usepackage{subcaption}
\usepackage{float} 
\usepackage{placeins} 
\usepackage{graphicx} 
\usepackage{amsmath, amssymb, amsfonts} 
\usepackage{cite} 
\IEEEoverridecommandlockouts 

\usepackage{algpseudocode}
\SetCommentSty{mycommfont}
\SetKwInput{KwInput}{Input}                
\SetKwInput{KwOutput}{Output}     

\usepackage{textcomp}
\usepackage{xcolor}
\def\BibTeX{{\rm B\kern-.05em{\sc i\kern-.025em b}\kern-.08em
    T\kern-.1667em\lower.7ex\hbox{E}\kern-.125emX}}

\begin{document}

\title{Federated Learning Framework via Distributed Mutual Learning}

\author{\IEEEauthorblockN{1\textsuperscript{st} Yash Gupta}
\IEEEauthorblockA{\textit{Computer Science Department)} \\
\textit{Lakehead University}\\
Thunder Bay,ON, Canada \\
ygupta1@lakeheadu.ca}
}

\maketitle

\begin{abstract}
Federated Learning has gained a significant amount of attraction in recent years. Federated learning enables clients to learn about their private data and then share their learnings with the central server to create a generalized global model and further share the generalized model with all clients. This aggregation of knowledge is based on the aggregation of model weights which has many associated issues. The model weights are more susceptible to model inversion attacks and would use a significant amount of bandwidth to share. In this work, we propose a loss-based federated learning framework using deep mutual learning between all clients using knowledge distillation. We use face mask detection as our case study to validate our work. We compared our results with traditional and asynchronous weight-updating federated learning models to show the advantage of the proposed framework, i.e., sharing losses instead of weights while maintaining high accuracy.
\end{abstract}


\section{Introduction}
Concerns regarding data privacy are at an all-time high. It is a public concern with rules enforced by laws such as General Data Protection Regulation (GDPR) in the European Union. Therefore data generated by IoT devices like smartphones and watches or data generated by public organizations like schools and hospitals cannot be gathered by one central entity, which poses a considerable problem for traditional machine learning. In traditional machine learning, data is collected by a central entity, which then trains a model and deploys it. Federated learning provides a solution to the problem mentioned above. In vanilla federated learning, multiple clients collect their data and train their respective models. After training their models, the models share the weights or their calculated gradients with a central authority, aggregating the weights or the shared gradients and distributing the average to the clients. There are several significant issues associated with the vanilla approach. Sharing the entire weight of the model leaves room for a model inversion attack which can reveal information about the sensitive data used during the training process.
\\
Sharing the entire weights of the models can also strain the network bandwidth. Since we are sharing the weights, the vanilla federated learning also makes a strong assumption of homogeneous client models, i.e., all the client models share the same architecture. This assumption is not ideal since different IoT devices have different computational abilities and might use different architectures. In this paper, we propose a loss-based federated learning framework that allows the clients to update their weights based on distributed mutual learning.
\\

The remainder of the paper is organized as follows. Section II provides the relevant background and literature review. In section III, we present the proposed framework i.e., federated learning using mutual learning via knowledge distillation. Our proposed method is evaluated in section IV. Section V comments on the observed results. Finally, the paper is concluded in section VI.
\section{Literature Review}
\begin{algorithm}[!h] 
\small 
\SetAlgoNlRelativeSize{-1} 
\SetInd{0.3em}{0.5em} 
\DontPrintSemicolon 

\caption{Federated Learning Using Distributed Mutual Learning}
\label{alg:FL-DML}

\SetKwInOut{Input}{Input}

\Input{\textbf{Clients}, \textbf{Rounds}}

Initialize Stratified K-Folds: \texttt{Fold $\leftarrow$ (1+Clients) × Rounds +1} \\     
Load Dataset \\
$G \leftarrow$ Global Model \\
$Clients \leftarrow$ List of Client Models \\
$ModelRes \leftarrow$ List of Client Model Performance Metrics \\
$G, GRes \leftarrow$ \texttt{genModel(Fold.pop(), G)} \\

\ForEach{$c \in Clients$} {
    $c.set\_weights \leftarrow G.get\_weights$
}

\For{$i \in Rounds$} {
    \For{$c \in Clients$} {
        $Clients[c], ModelRes[c] \leftarrow \texttt{genModel(Fold.pop(), Clients[c])}$
    }

    $Layer \leftarrow$ Shallow \\

    \If {$(i+1) \mod \delta == 0$ \textbf{and} $i \geq 5$} {
        $Layer \leftarrow$ Deep \\
    }

    $clientWeights \leftarrow$ \texttt{preprocessWeights(Clients, ModelRes)} \\
    $avgWeights \leftarrow$ \texttt{averageWeights(clientWeights)} \\

    $G.set\_weights \leftarrow$ \texttt{updateWeights(G.get\_weights, avgWeights, Layer)} \\
    $G, GRes \leftarrow$ \texttt{genModel(Fold.pop(), G)} \\

\ForEach{$c \in Clients$} {
    $c.set\_weights \leftarrow \texttt{updateWeights}(c.get\_weights,$ \\
    \hspace{2em} $\texttt{avgWeights, Layer)}$
}

}
\end{algorithm}
The literature review is divided into weight-based or traditional federated learning methods and loss-based federated learning methods with a background in knowledge distillation and mutual learning.
\subsection{Weight Based Federated Learning}
Federated learning was first introduced by Google \cite{google} as a means to develop a decentralized system for training machine learning models on private data and to aggregate the learnings of different models to generalize the distributed models over time. Traditional federated learning trains the models on local data and aggregates the weights of the distributed models, and averages them. The averaged weights are then shared with the models. Sharing the entire weight of the model leaves room for a model inversion attack which can reveal information about the sensitive data used during the training process. Sharing the entire weights of the models can also strain the network bandwidth. 
\\
To overcome the problem mentioned above, researchers proposed asynchronous weight-updating federated learning methods. In this framework, we only share the weights of a part of the model instead of sharing the entire weight. For instance, Sakib et al. \cite{sakib} employed an asynchronous weight updating federated learning for ECG (electrocardiograph) analytics for arrhythmia detection. Sakib et al. [11] demonstrated a comparison between synchronous and asynchronous weight updating methods and indicated that with the increasing number of clients, the asynchronous weight updating method provides encouraging performance. Zhang et al. \cite{zhang} adopted a different approach for the same problem by introducing a distributed arrhythmia detection algorithm based on ECG data to allow collaboration between caregiving institutions. They applied an Elastic Weight Consolidation (EWC) algorithm to allow the federated learning models to converge. They also trained the global model on a global dataset with the aggregated local weights to minimize the global error. In our previous work \cite{yash}, we used methodologies to combine weight updating of models asynchronously \cite{chen}–\cite{zubair} and global error minimization on a global data split during every round. \cite{yash} performed a weighted average by using scoring metrics like accuracy to weigh the model weights during aggregation.
\begin{figure*}[t!] 
	\centering
	\includegraphics[width=\textwidth]{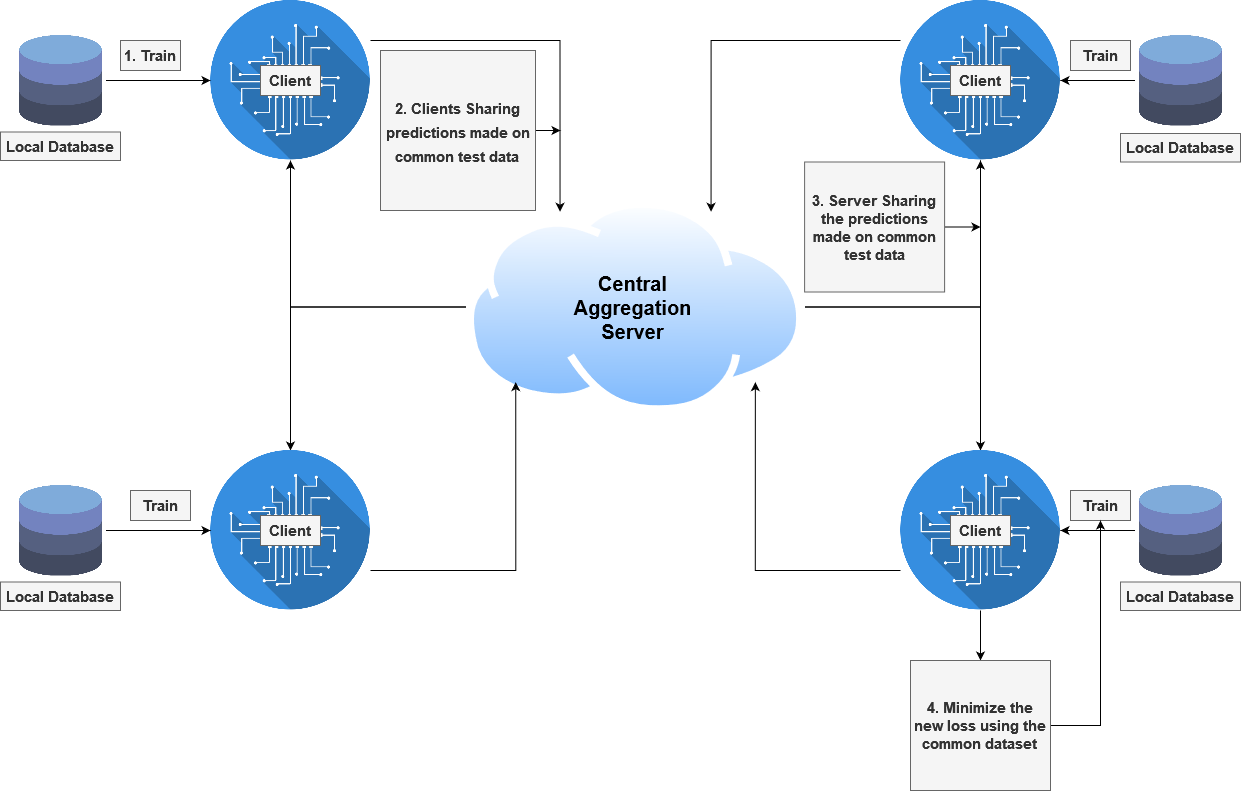} 
	\caption{Proposed model architecture.}
    \label{modelarch}
\end{figure*}
\subsection{Loss Based Federated Learning}
Even though sharing weights can help achieve good results, it has shortcomings in terms of data privacy and bandwidth usage. To combat the problems mentioned above, researchers have been interested in sharing the losses of the models to allow the models to learn from each other at a global level. This has been achieved using knowledge distillation and mutual learning. 
\\
Hinton, Vinyals, and Dean first proposed the idea of distilling knowledge in a neural network \cite{hinton}. The idea behind this approach is to train a large network and many special models. The large network is a general model that can classify a large number of classes, whereas the special models are specialized in classes that the large network fails to classify properly. The knowledge of the two different kinds of models is then distilled into a smaller model for shorter inference time and smaller size. This model is supposed to learn the best of the large network as well as the special models and is able to classify all the classes well which was not the case for the large network and special models. The distillation is perfromed by using a KL (Kullback Leiber) divergence loss which is a measure of how a probability distribution differs from another probability distribution.
\\
Based on the concept of knowledge distillation, Ying Zhang \cite{ying} proposed deep mutual learning where the models in a network can learn from each other using the KL divergence loss where each model has the model loss (for ex. Binary Cross Entropy) and the KL difference loss which is calculated using the predictions from the other models in the network. The idea behind the approach is that over time the models will mimic each other and will fulfill each other's gaps over time.
\\
Lukman and Yang further improve upon the ideas of Hinton and Zhang by improving deep mutual learning via knowledge distillation \cite{lukman}. Lukman and Yang adopt a teacher network and distill its knowledge in a mutual learning setting proposed by \cite{ying}. The loss function comprises the model loss, an average KL divergence loss based on other models in mutual learning, and another KL divergence loss between the model and the teacher model.
\\
Based on the concepts explained before, researchers have proposed multiple federated learning frameworks which allow the distributed models to learn from each other like the models in mutual learning. Li and Wang proposed FedMD, which is a heterogeneous federated learning framework using model distillation \cite{li}. The framework allows the clients to train models on their local data and then provides a common global dataset on which all clients make their predictions and compute the losses. The losses are aggregated by the server, which calculates an average loss that is transmitted to each model. The models then optimize their models based on the new average loss using the common public dataset. Shen et al. proposed a federated mutual learning framework \cite{shen}. In the federated mutual learning framework, the server keeps a global model, whereas the client has a meme model as well as a local model. The meme model is a copy of the global model at the time of initialization. The client trains the meme model and local model using deep mutual learning. The server aggregates the meme models and takes the average of all the meme model weights to generalize a global model. This allows the client to have a local model as well as generalize a global model.
\\
In this paper, we propose a distributed mutual learning federated learning framework in which the models share the losses calculated on a common test set and perform deep mutual learning at every round of training. 
\begin{table}[h!]
\centering
\caption{Dataset class count.}
{\renewcommand{\arraystretch}{1.2}%
\begin{tabular}{|l|l|l|}
\hline
\textbf{Dataset}   & \textbf{Mask} & \textbf{No Mask} \\ \hline
\textbf{Dataset 1} & 1915          & 1918             \\ \hline
\textbf{Dataset 2} & 2994          & 2994             \\ \hline
\end{tabular}}
\label{tab-datasets}
\end{table}
\section{Methodology}
This section describes the proposed framework architecture, the experimental setup to validate the architecture, the case study validate the architecture and the test environment.
\begin{figure*}[t!] 
	\centering
	\includegraphics[width=\textwidth]{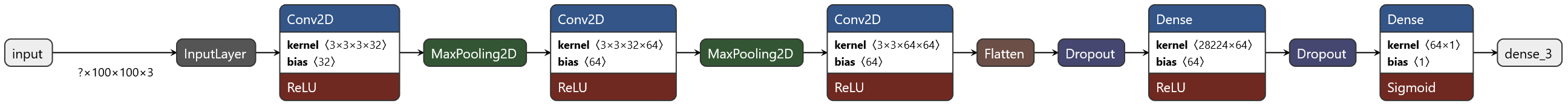} 
	\caption{Proposed model architecture.}
    \label{modelarch}
\end{figure*}
\begin{table*}[]
\centering
\caption{Federated Learning Framework Performance on Dataset 2}
\begin{tabular}{l|l|l|l|l|l|}
\cline{2-6}
                                                                               & \textbf{Client 0} & \textbf{Client 1} & \textbf{Client 2} & \textbf{Client 3} & \textbf{Client 4} \\ \hline
\multicolumn{1}{|l|}{\textbf{Vanilla Federated Learning}}                      & 92.65             & 92.65             & 92.65             & 92.65             & 92.65             \\ \hline
\multicolumn{1}{|l|}{\textbf{Asynchronous Weight Updating Federated Learning}} & 93.7              & 91.42             & 94.49             & 93.30             & 90.78             \\ \hline
\multicolumn{1}{|l|}{\textbf{Mutual Learning Federated Learning}}              & 94.89             & 94.27             & 94.42             & 94.44             & 94.21             \\ \hline
\end{tabular}
\label{testtable}
\end{table*}
\begin{figure*}[!h] 
	\centering
	\includegraphics[width=1\textwidth]{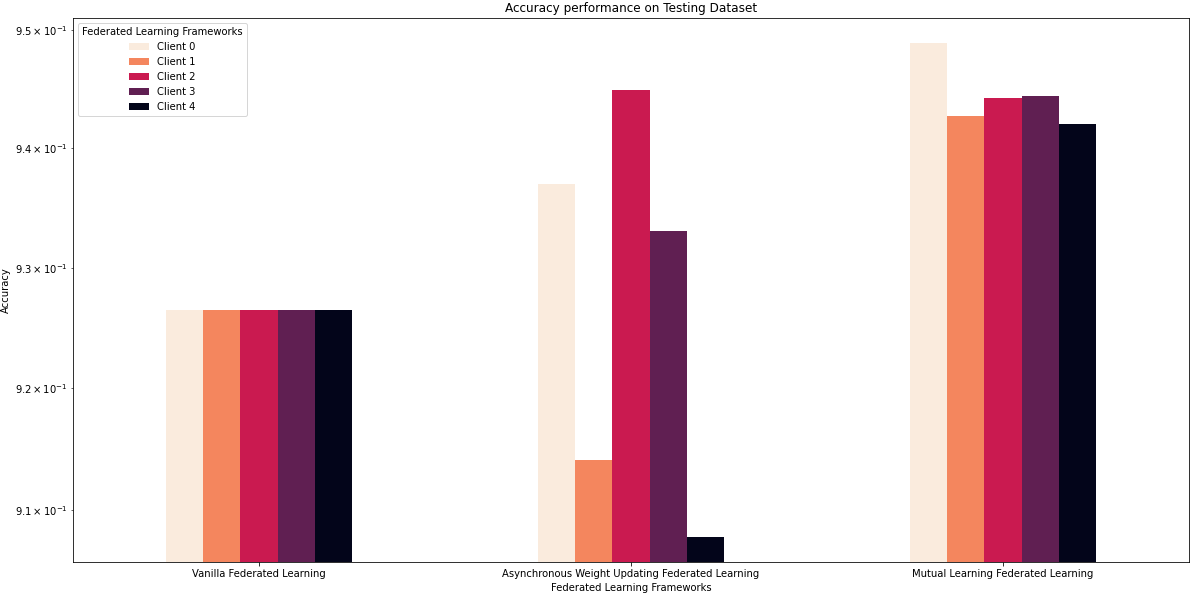} 
	\caption{Client Performances on Testing Data}
    \label{testing}
\end{figure*}

\subsection{Proposed Architecture}
The proposed federated learning framework is outlined in Algorithm~\ref{alg:FL-DML}, which details the training and update process within our distributed mutual learning-based approach.

Our literature review identified a research gap: existing federated learning frameworks do not leverage distributed mutual learning, relying instead on direct weight aggregation. To address this, we propose a federated learning framework that enables mutual knowledge sharing among clients without weight transmission.

As described in Algorithm~\ref{alg:FL-DML}, the framework begins by initializing a global model using publicly available data. Clients can either adopt the global model’s initial weights or initialize their own models using the public dataset. Each client then trains on its own local dataset, optimizing its model based on individual data distributions.

After completing the local training round, clients perform inference on a dynamically changing test dataset provided by the central server. This common test dataset varies in each round, exposing models to diverse examples and enhancing generalization. The server aggregates results from all clients and redistributes them, allowing each client to refine its model based on the collective knowledge of the system.

Instead of traditional gradient or weight sharing, clients update their models using a loss-based optimization strategy as formulated in Equation~\ref{eq:1}:

\begin{equation}\label{eq:1} Loss = Model_{loss} + KLD_{avg} \end{equation} \begin{equation} \label{eq:2} KLD_{avg} = \frac{1}{K-1}\sum_{j=1,j!=i}^{K} P_{i} \log\frac{P_{i}}{P_{j}} \end{equation}

Equation~\ref{eq:2} defines the average KL divergence loss, where ii represents the local model, jj represents the other client models, and KK is the total number of clients. The intuition behind this approach is that, over time, models will gradually converge and mimic each other, benefiting from deep mutual learning in a distributed environment.

By following the structured training, evaluation, and knowledge-sharing steps defined in Algorithm~\ref{alg:FL-DML}, this framework ensures privacy preservation, bandwidth efficiency, and improved model generalization while addressing security risks in traditional federated learning.

\begin{figure*}[h!]
\centering
\begin{subfigure}{\textwidth}
    \includegraphics[width=\textwidth]{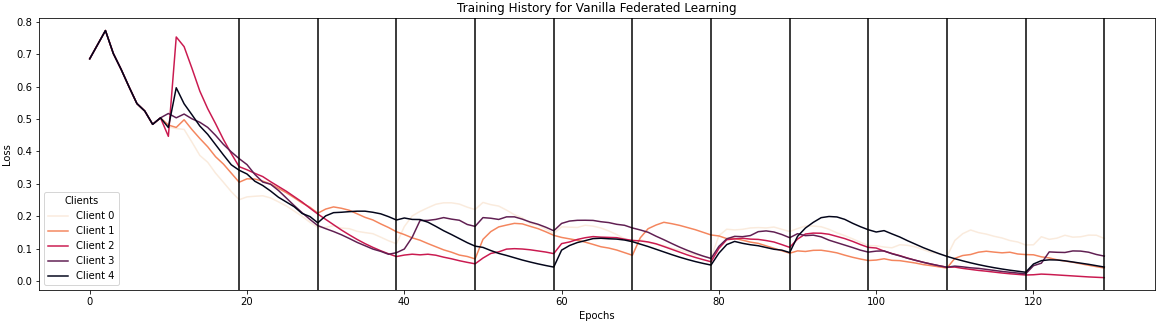}
    \caption{Synchronous Federated Learning Training History}
    \label{fig:first}
\end{subfigure}
\hfill
\begin{subfigure}{\textwidth}
    \includegraphics[width=\textwidth]{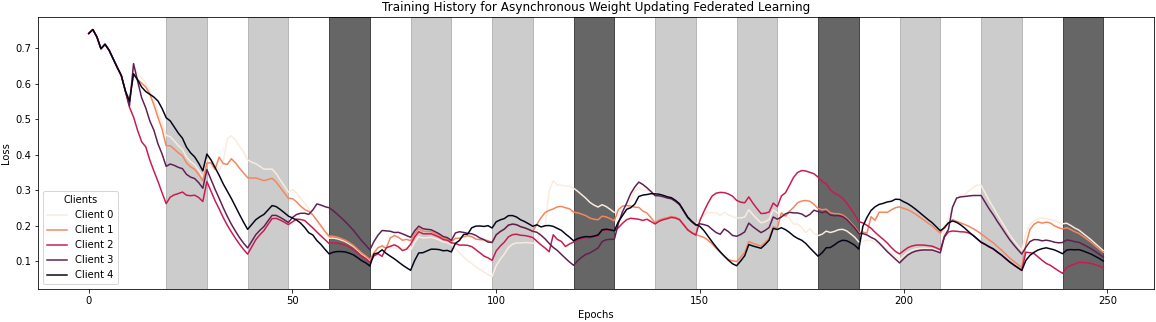}
    \caption{Asynchronous Weight Updating Federated Learning Training History}
    \label{fig:second}
\end{subfigure}
\hfill
\begin{subfigure}{\textwidth}
    \includegraphics[width=\textwidth]{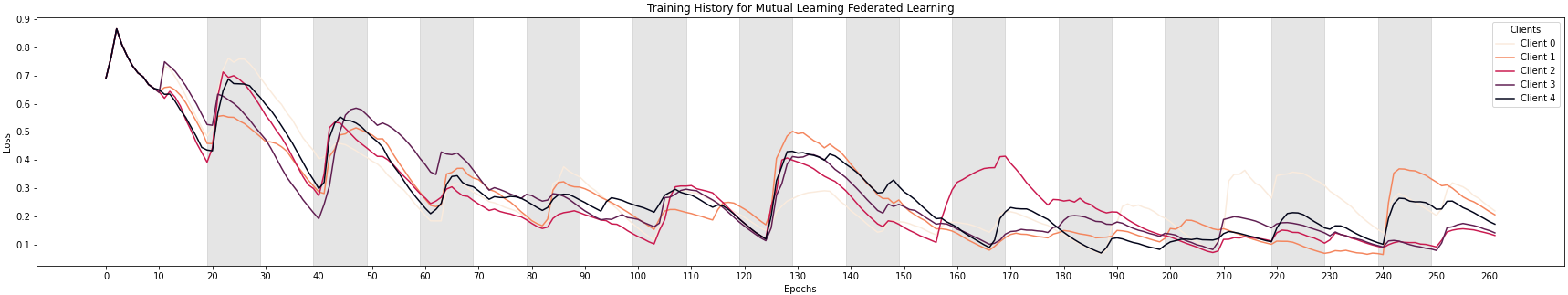}
    \caption{Distributed Mutual Learning Federated Learning Training History}
    \label{fig:third}
\end{subfigure}
\caption{Training History}
\label{trainfig}
\end{figure*}
\subsection{Experimental Setup}
\subsubsection{Dataset}
In order to validate our proposed framework we are using a face mask detection dataset which is a binary classification problem. Two facemask datasets were collected from public Github repositories~\cite{d1}  
and Kaggle repositories~\cite{d2}  
repositories, respectively. The dataset class counts are listed in Table~\ref{tab-datasets}. The first dataset is employed to train the global and client models while the second dataset is used to test the trained models in order to test the generalizability of the model. 

Since the data are collected from different sources, the following steps were performed to preprocess the data: 
\begin{itemize}
\item Resize image : $100\times100\times3$,
\item Normalize the image,
\item Convert the image to an array.
\end{itemize}
\subsubsection{Model Architecture}

In order to keep the model size small while keeping the performance relatively competitive, we are using a customized CNN model based on VisionNet \cite{yash}. The model architecture is presented in Fig.~\ref{modelarch}. The input size of the model is 100x100x3. VisionNet consists of three convolutional layers, where 2x2 max-pooling layers follow the first two convolutional layers. The last convolutional layer is followed by a dropout layer connected to a dense layer of 64 neurons, followed by another dropout layer that uses a sigmoid function for activation, i.e., treating the problem as a binary classification task using only half of the connections compared to softmax, i.e., treating the problem as a two-class problem. We added the two dropout layers to prevent overfitting.

\subsubsection{Testing Environment}
In order to validate our method, we are comparing it with Vanilla Federated Learning and Asynchronous Weight Updating federated learning. All frameworks are exposed to the same conditions:
\begin{itemize}
    \item Same model architecture
    \item Same data size for each training round
    \item Same number of epochs
    \item IID dataset
\end{itemize}
\section{Results}

Figure \ref{trainfig} shows the model training loss history over the period of 12 rounds. The asynchronous weight updating method, uses a delta of 3 i.e. share the deep weights every 3rd round. The shaded regions as well as the horizontal lines on the graphs represents the time or epoch when we shared the weights or performed collaborative learning. In asynchronous weight updating training graph, the lighter shade meaning the time when we share the shallow weights whereas the dark shades means we are sharing the deep weights. Figure \ref{testing} and Table \ref{testtable} represents the client accuracies trained on different frameworks on a completely unseen dataset i.e. dataset 2.
\section{Discussion}
From the results, we can see the training history of all the federated learning models. We can see that out of all three methods. The vanilla federated learning converges the fastest since we are sharing the entire weight. Since we are sharing the weights asynchronously, we do observe more peaks during the weight-sharing and optimization process. In the proposed framework, we can observe a downward trend, i.e., The models are converging to a minimum loss. Moreover, we can see in the shaded region, that is where the KL divergence loss is taking place. These spikes in the loss are described by the additional KL divergence loss and, as we can see, also follows a downward trend as we progress through the training. Moreover, we can observe that over time the clients do mimic each other and hence generate more generalized models. This theory can further be confirmed by \ref{testtable} and \ref{testing}. We can see that the proposed framework produces more generalized models with almost the same performance, which cannot be said for the asynchronous weight updating models. Moreover, since our models were exposed to more data during the common test phase, they learned more overall compared to the synchronous method.
\section{Conclusion \& Future Works}
In this work, we proposed a federated learning framework based on distributed mutual learning, which uses deep mutual learning at its core to produce more generalized models. In our work, we compared our framework with weight-sharing methods and outperformed them. This is a novel contribution since we are only sharing the common losses on new public data, i.e., new data for every round collected by the central server, which helps the model to learn from more data. In this paper, we only used a framework to solve the problem of computer vision and assumed IID data and homogeneous models. In future iterations of the project, we would like to explore more real-world problems, the effects of different model architectures on each other, and non-IID data. We would also like to explore knowledge distillation through a large global model.
\section{Acknowledgement}
The research was performed by a single author, therefore all the work was done by the author.

\end{document}